\documentclass[cameraready]{template/mac_automl_xmu_blue}

\usepackage{array}
\usepackage{tabularx}
\usepackage{float}

\newcommand{\internshipmark}{\ensuremath{\dagger}}
\newcommand{\projectleadermark}{\ensuremath{\ddagger}}
\newcommand{\oursrow}{\rowcolor{omniaccent!7}}
\newlength{\introfigureimageheight}
\newlength{\qualitativeimageheight}

\setleftheadercontent{%
  \headerlogo[-0.06\height]{10.8mm}{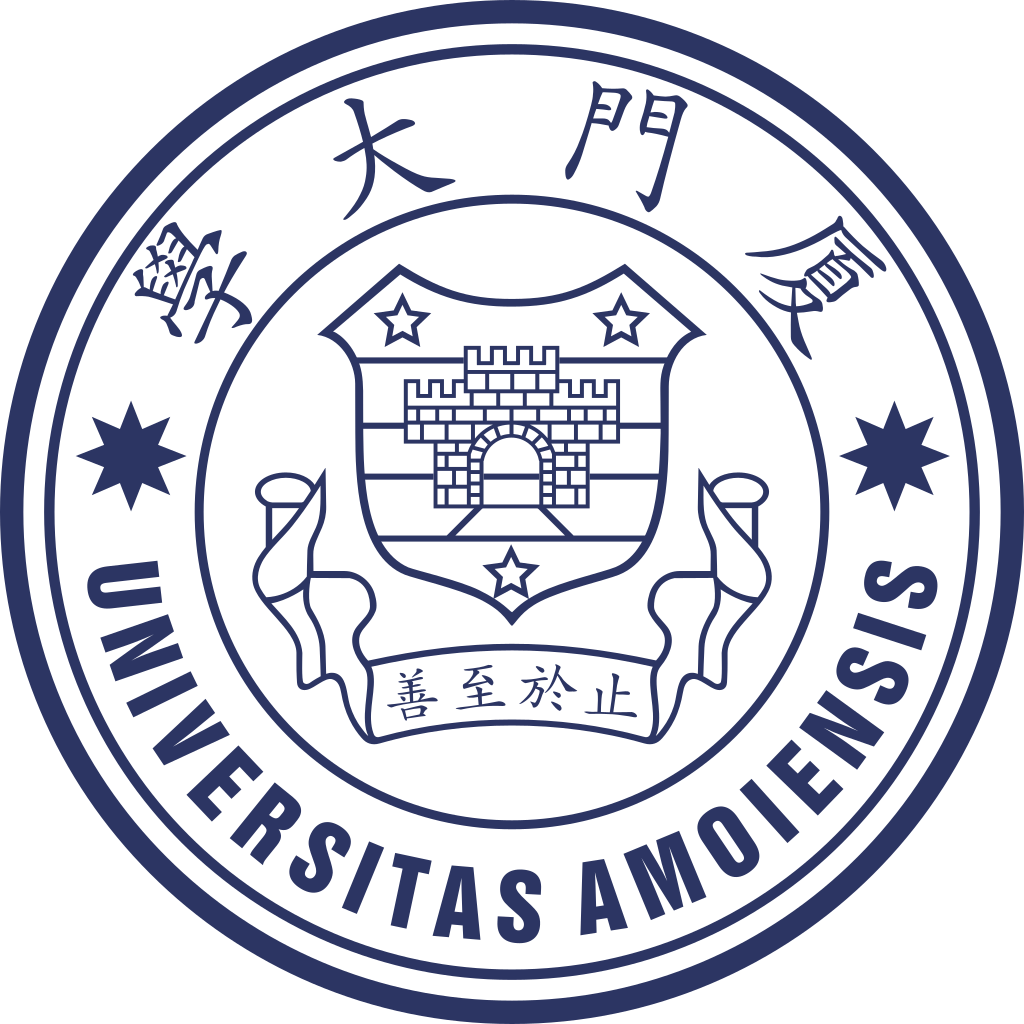}%
  \headerlogospace{3.2mm}%
  \headerlogo[-0.06\height]{10.8mm}{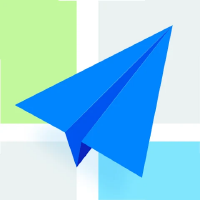}%
  \headerlogospace{3.2mm}%
  \headerlogo[-0.06\height]{10.8mm}{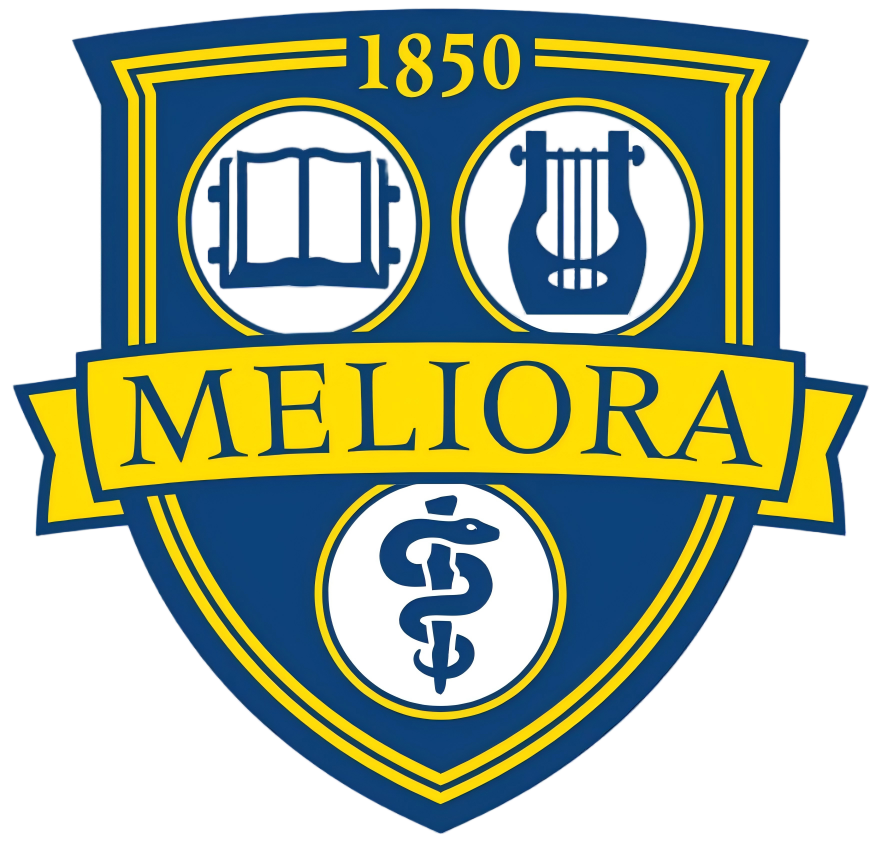}%
}
\setrightheadericon{%
  \raisebox{-0.06\height}{\includegraphics[height=10.8mm,trim=0 24bp 0 32bp,clip]{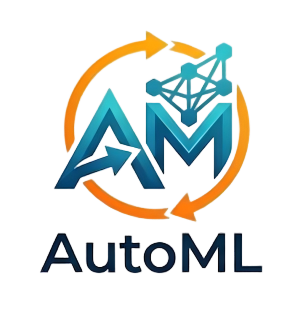}}%
}
\setrunningheadericon{%
  \headerlogo[-0.03\height]{8.2mm}{assets/branding/automl.png}%
}
\setheadergroupname{MAC-AutoML}

\title{\textcolor{omniaccent!88!black}{One Ranking, Any Budget:} Matryoshka Evidence-to-Context\\[-0.4mm]
Frame Selection for Long-Video Understanding}
\setfrontauthors{%
  \authorrow{%
    \authorentry{\href{https://openreview.net/profile?id=~Wang_Chen10}{Wang Chen}}{1,2,\internshipmark},\authorsep
    \authorentry{\href{https://openreview.net/profile?id=~Yu_Chen85}{Yu Chen}}{2},\authorsep
    \authorentry{\href{https://openreview.net/profile?id=~Xiang_Wang2}{Xiang Wang}}{2,\projectleadermark},\authorsep
    \authorentry{\href{https://openreview.net/profile?id=~Shuai_Li46}{Shuai Li}}{2},\authorsep
    \authorentry{\href{https://openreview.net/profile?id=~Jinfa_Huang2}{Jinfa Huang}}{3},\authorsep
    \authorentry{\href{https://openreview.net/profile?id=~Xiawu_Zheng1}{Xiawu Zheng}}{1,\corremailmark}%
  }%
}
\setfrontaffiliations{%
  \affiliationline{\authormark{1}Media Analytics and Computing Lab, Xiamen University, Xiamen, China}
  \affiliationline{\authormark{2}Amap, Alibaba Group, China}
  \affiliationline{\authormark{3}Department of Computer Science, University of Rochester, Rochester, NY, USA}
}
\setfrontcontact{%
  \contactline{{\bfseries \projectleadermark\ Project Leader \qquad \corremailmark\ Corresponding Author}}
  \contactline{\vspace{0.7mm}{\fontsize{7.6}{9.2}\selectfont\color{black!58}\textsuperscript{\internshipmark}\ This work was conducted during the authors{\rmfamily\textquotesingle} internships at Amap, Alibaba Group.}}
}
\usecustomauthorlayout

\hypersetup{
  pdftitle={One Ranking, Any Budget: Matryoshka Evidence-to-Context Frame Selection for Long-Video Understanding},
  pdfauthor={Wang Chen, Yu Chen, Xiang Wang, Shuai Li, Jinfa Huang, Xiawu Zheng}
}

\abstract{
Frame selection is essential for applying Large Multimodal Models (LMMs) to long videos due to severe frame redundancy and limited context windows. Since the appropriate frame budget varies with the downstream LMM, reasoning demands, and latency constraints, a practical selector should serve multiple budgets. However, existing methods typically optimize an isolated frame subset for each predefined budget: when the budget changes, previously selected evidence may be replaced rather than progressively augmented. 
A fixed-weight ranking allows prefix reuse across budgets but applies the same weighting at every position, overlooking the distinct roles of early and later ranks.
We formulate long-video frame selection as a Matryoshka ranking problem: constructing a single priority sequence whose small prefixes concentrate query-conditioned evidence, while progressively larger prefixes preserve this evidence and add broader temporal context. Efficiently constructing such a ranking is itself challenging, as densely sampling long videos and evaluating frame–query relevance incurs substantial overhead. We therefore introduce \textbf{M}atryoshka \textbf{E}vidence-to-\textbf{C}ontext (\textbf{MEC}) Frame Selection, a training-free framework that builds a reusable sparse video index, discovers candidates through sparse probing and local zooming, and greedily constructs a position-adaptive ranking—early positions emphasize evidence; later positions progressively favor temporal coverage while preserving visual diversity. A single ranking can thus be truncated to any target budget without rerunning the selector. Across four benchmarks and six frame budgets, \textbf{MEC improves average accuracy over uniform sampling by 3.77 points, matches strong state-of-the-art selectors, and reduces end-to-end selection latency by 47.37-51.19\%.}
}

\begin{document}
\maketitle

\section{Introduction}

\begin{figure}[t]
\centering
\captionsetup[subfigure]{font=small,labelfont=bf,justification=centering,singlelinecheck=true}
\setlength{\introfigureimageheight}{0.436\columnwidth}
\begin{subfigure}[t]{0.52\columnwidth}
  \centering
  \begin{minipage}[c][\introfigureimageheight][c]{\linewidth}
    \centering
    \includegraphics[width=\linewidth]{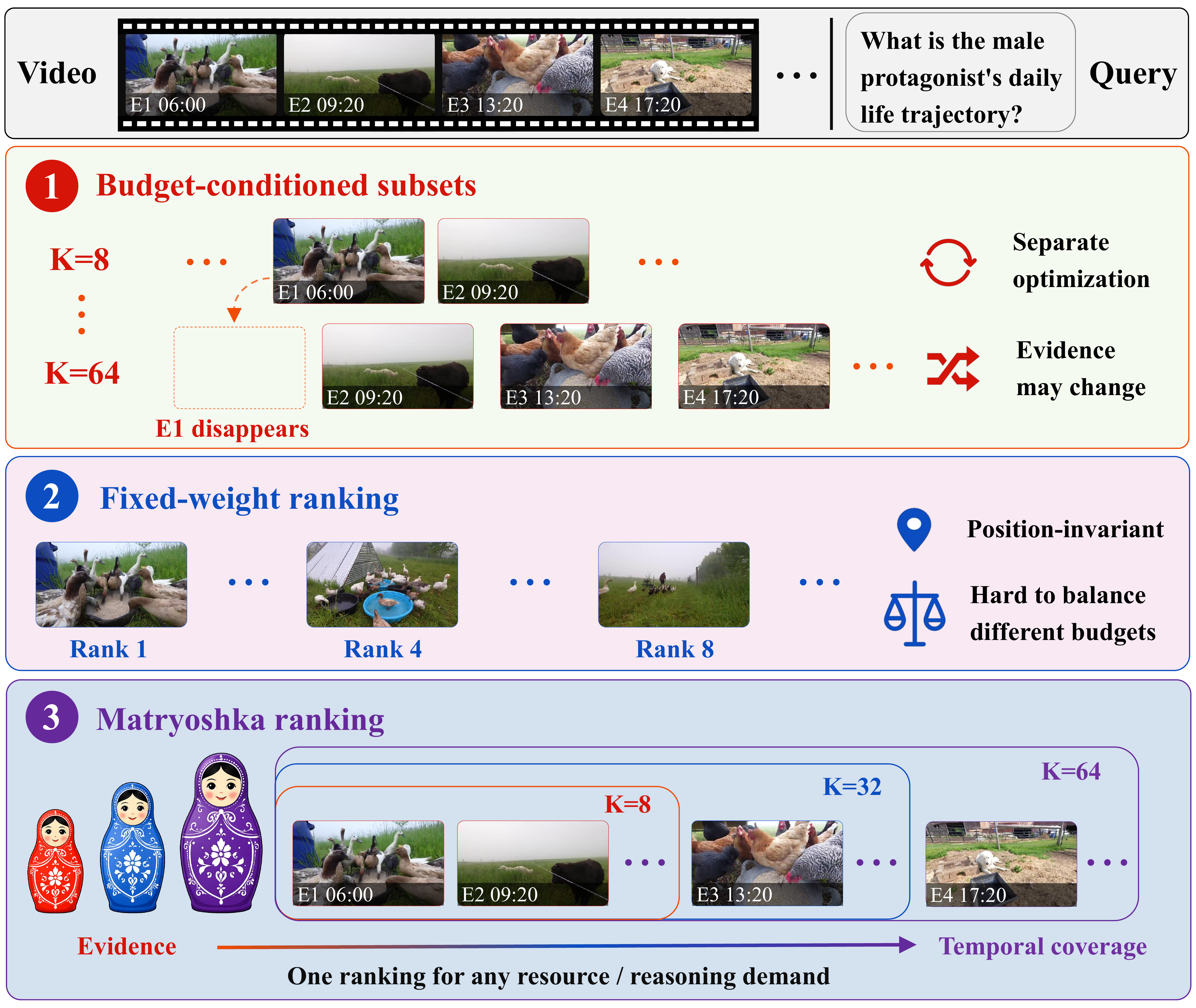}
  \end{minipage}
  \caption{\textbf{Frame-selection paradigms}}
  \label{fig:motivation}
\end{subfigure}\hfill
\begin{subfigure}[t]{0.45\columnwidth}
  \centering
  \begin{minipage}[c][\introfigureimageheight][c]{\linewidth}
    \centering
    \includegraphics[width=\linewidth]{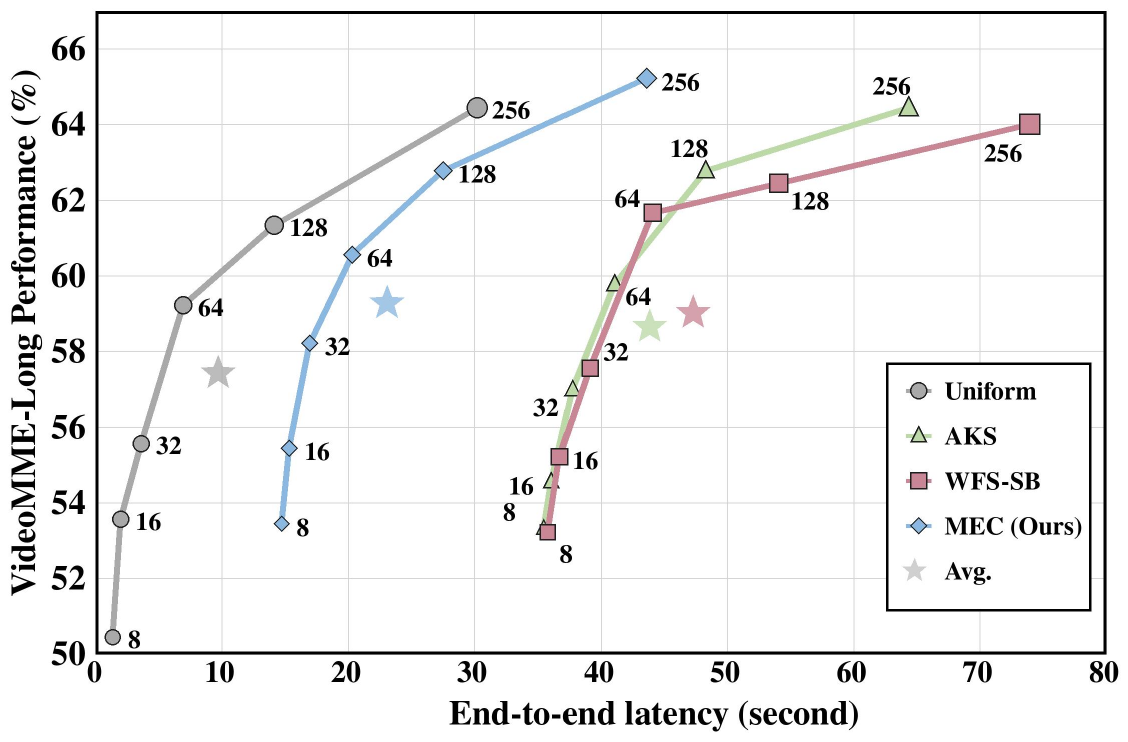}
  \end{minipage}
  \caption{\textbf{Accuracy--latency trade-off}}
  \label{fig:overview}
\end{subfigure}
\caption{Motivation and empirical trade-off of \textsc{MEC}. (a) Budget-conditioned selection optimizes each subset separately and may replace earlier evidence; fixed-weight ranking reuses prefixes but struggles to balance performance across budgets. \textsc{MEC} instead constructs one Matryoshka ranking whose prefixes preserve early evidence and progressively extend temporal context. (b) On Video-MME Long, each marker is annotated with its frame budget and stars denote six-budget averages. \textsc{MEC} provides a favorable accuracy--latency balance relative to stronger budget-specific selectors.}
\label{fig:intro-overview}
\vspace{-3pt}
\end{figure}

Recent advances in Large Multimodal Models (LMMs) \citep{bai2025qwen3vl,wang2025internvl35,clark2026molmo2,an2026llavaov2} have extended multimodal reasoning from static images to video. Applications include video question answering, temporal event localization, dense captioning, and video object tracking \citep{shaar2026movierecapsqa,zhang2026timelens,fiastre2026captionformer}. Yet scaling from short clips to long-form video remains challenging. A long recording may contain tens of thousands of highly redundant frames, while the evidence needed to answer a question may be confined to a fleeting action, a subtle state transition, or several events separated by minutes.

Processing every frame is infeasible, yet indiscriminate temporal subsampling can miss brief, subtle, or temporally distant evidence. Keyframe selection retains a compact set of query-relevant frames for the downstream LMM. Agentic or iterative temporal search can adaptively locate evidence \citep{zhang2025dvd,zou2026air}, but repeated decisions, VLM evaluations, and, where applicable, tool invocations introduce additional search steps and latency. Visual-token compression shortens token sequences through hierarchical aggregation, merging, or pruning \citep{li2026videochatflash,fu2025framefusion,alvar2025divprune}, but aggressive compression may discard fine-grained, query-critical evidence. Frame selection reduces redundant input while preserving visual evidence at its original fidelity. What should a frame-selection strategy provide for long-video understanding?

A selection strategy should remain effective as the frame budget changes. Many existing keyframe methods optimize a separate subset for each predefined budget $K$. Because the appropriate budget varies with the downstream LMM, reasoning demands, and latency constraints, changing $K$ requires rerunning the selector. Moreover, as illustrated in Figure~\ref{fig:motivation}, evidence selected under a tight budget may disappear as the budget increases. A fixed-weight ranking can serve different budgets through nested prefixes, but it uses the same balance of evidence, temporal coverage, and visual diversity at every rank, limiting its ability to concentrate evidence in short prefixes and expand context in longer ones. Inspired by the Matryoshka principle explored in representation learning and multimodal visual-token modeling \citep{kusupati2022matryoshka,cai2025matryoshka}, we formulate long-video frame selection as the \emph{Matryoshka ranking problem}: the task of constructing one budget-independent priority sequence whose short prefixes concentrate query-conditioned evidence, while progressively longer prefixes preserve that evidence and add complementary context.

Constructing such a ranking efficiently poses a challenge. Densely sampling a long video and evaluating each frame against the query can incur substantial selection cost. We address this challenge with Matryoshka Evidence-to-Context (\textsc{MEC}) Frame Selection, a training-free framework that separates reusable video analysis from query-conditioned discovery. \textsc{MEC} first builds a sparse index of low-resolution appearance, local visual change, and observability. For each query, it scores sparse probes using multiple cues, activates high-value segments, and zooms around anchors to form a compact candidate pool. It then greedily constructs the ranking using a position-dependent marginal surrogate: early positions emphasize query-conditioned evidence, whereas later positions progressively increase temporal coverage while maintaining visual diversity. Any target budget can be served by truncating this ranking and restoring chronological order, without rerunning the selector. Across four long-video benchmarks and six frame budgets, \textsc{MEC} remains competitive with strong methods while offering a favorable accuracy--latency trade-off. As shown in Figure~\ref{fig:overview}, \textbf{\textsc{MEC}'s six-budget average on the Long subset of Video-MME exceeds AKS by $0.63$ accuracy points while reducing cached end-to-end average latency by $47.37\%$}.

Our contributions are as follows: \textbf{1)} We formulate the Matryoshka ranking problem for long-video frame selection: one sequence serves all budgets, concentrating query-conditioned evidence in short prefixes and progressively expanding context in longer ones. \textbf{2)} We introduce \textsc{MEC}, a training-free framework that efficiently constructs a position-adaptive Matryoshka ranking to serve multiple frame budgets in a single run. \textbf{3)} Extensive experiments across four long-video benchmarks and six frame budgets demonstrate the effectiveness of \textsc{MEC} and its favorable accuracy--latency trade-off.

\section{Related Work}
\vspace{-5pt}

\noindent\textbf{Frame Selection for Long-Video Understanding.}
Frame selection fits long videos into the limited context windows of LMMs. Existing methods fall into several overlapping categories. Query-aware sampling prioritizes relevant frames, optionally balancing coverage or diversity, as in AKS, Nar-KFC, BOLT, MDP$^3$, and AdaRD-Key \citep{tang2025aks,fang2026narkfc,liu2025bolt,sun2025mdp3,zhang2025adardkey}. Structure-aware selection partitions or hierarchically organizes a video into clips, events, or scenes before allocating or refining frames, as in EFS, WFS-SB, and InfoShot \citep{chen2026efs,chen2026wfssb,zhao2026infoshot}; DIG further adapts selection to global or localized queries \citep{li2026dig}. Temporal search probes promising regions or iteratively acquires evidence, as in AKeyS, A.I.R., FOCUS, and FrameThinker \citep{fan2025akeys,zou2026air,zhu2026focus,he2026framethinker}. Learned selectors use model-generated supervision or reinforcement learning, including the M-LLM-based selector of \citet{hu2025mllmselection}, the RL-based selector of \citet{qin2026rlselection}, and ReaSon \citep{zhou2026reason}. Frame/token co-optimization determines which temporal content to retain and how many visual tokens or what resolution to assign to it, as in Q-Frame, KVTP, and QuoTA \citep{zhang2025qframe,liu2025kvtp,luo2025quota}. These categories overlap, emphasizing subset quality, search decisions, or token allocation. In contrast, \textsc{MEC} constructs a single reusable nested sequence using a position-dependent marginal surrogate, concentrating evidence in early ranks while progressively expanding temporal context in later ranks. Section~\ref{sec:appendix-related-work} further discusses video large multimodal models, Matryoshka representations, and visual-token compression.

\begin{figure}[t]
\centering
\includegraphics[width=0.99\textwidth]{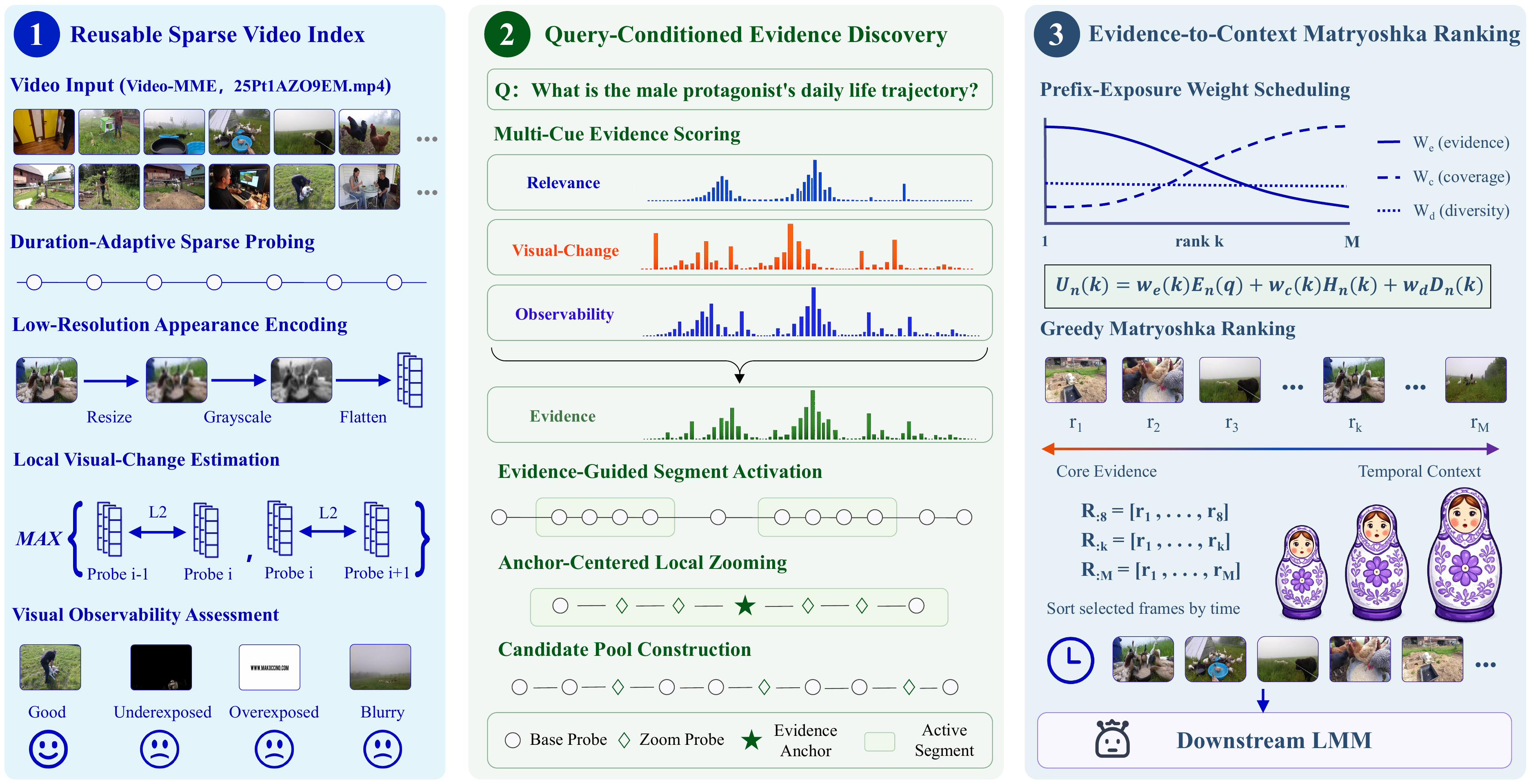}
\caption{Overview of \textsc{MEC}. A reusable sparse video index caches low-resolution appearance encodings, local visual-change estimates, and observability priors. For each query, \textsc{MEC} scores multi-cue evidence, activates promising segments, and performs local zooming before constructing a position-adaptive Matryoshka ranking whose nested prefixes support any frame budget.}
\label{fig:method}
\vspace{-10pt}
\end{figure}
\vspace{-5pt}

\section{Method}
\vspace{-8pt}

Figure~\ref{fig:method} summarizes \textsc{MEC}'s three stages: building a reusable sparse index, discovering query-conditioned evidence, and constructing an evidence-to-context Matryoshka ranking.

\subsection{Problem Formulation}

Let $\mathcal{V}$=$\{x_1,\ldots,x_N\}$ be a video, where $x_n$ is its $n$-th frame and $N$ is the total number of frames, and let $q$ be a question about the video. Given a frame budget $K$, a predefined-budget selector returns
\begin{equation}
\mathcal{F}_{K}=\Phi(\mathcal{V},q,K),
\qquad |\mathcal{F}_{K}|=K,
\end{equation}
where $\mathcal{F}_{K}$ is optimized for $K$, so changing the budget requires reevaluating $\Phi$. \textsc{MEC} instead invokes its selector once with a budget-independent sequence length $1\leq M\leq N$:
\begin{equation}
\begin{aligned}
\mathcal{R}&=\Psi(\mathcal{V},q,M)=(r_1,r_2,\ldots,r_M),\\
\mathcal{F}^{\mathrm{MEC}}_{K}&=\{r_1,\ldots,r_K\},\qquad 1\leq K\leq M.
\end{aligned}
\end{equation}
Here $r_k$ is the original-frame index at rank $k$, and all $r_k$ are distinct. The selected sets are nested across budgets, but nesting alone does not ensure useful prefixes. Let $\mathcal{B}\subseteq\{1,\ldots,M\}$ be the supported budgets and $\pi_K\geq0$ their deployment importance, with $\sum_{K\in\mathcal{B}}\pi_K=1$. We use $\mathcal{Q}_K(\mathcal{R}_{:K};\mathcal{V},q)$ to denote the ideal downstream utility of prefix $\mathcal{R}_{:K}=(r_1,\ldots,r_K)$. This oracle utility is unavailable: correctness is unknown at inference time, and repeatedly evaluating candidate prefixes with the LMM is prohibitively expensive. The ideal Matryoshka objective is
\begin{equation}
\mathcal{R}^{\star}\in\arg\max_{\mathcal{R}\in\Pi_M}
\sum_{K\in\mathcal{B}}\pi_K\,
\mathcal{Q}_K(\mathcal{R}_{:K};\mathcal{V},q),
\label{eq:matryoshka-objective}
\end{equation}
where $\Pi_M$ contains all sequences of $M$ distinct frame indices. Equation~\eqref{eq:matryoshka-objective} specifies the desired all-prefix ranking, not a directly computable selector loss. A frame at rank $k$ appears in every supported prefix with $K\geq k$, giving cross-budget exposure $\alpha_k=\sum_{K\in\mathcal{B},\,K\geq k}\pi_K$ with $\alpha_1\geq\cdots\geq\alpha_M$. Early positions therefore affect a broader range of budgets. This positional asymmetry motivates prioritizing evidence useful under both tight and large budgets before expanding temporal and visual context. Section~3.4 implements this structure through a tractable marginal surrogate.

\subsection{Reusable Sparse Video Index}

Analyzing all $N$ frames for each question repeats video-level computation. \textsc{MEC} caches appearance, change, and observability cues in a sparse index independent of $q$.

\noindent\textbf{Duration-Adaptive Sparse Probing.}
For duration $T$ in seconds, base probe count $P_0$, maximum increment $P_1$, and saturation horizon $\tau_p$, we use the uniform probe count
\begin{equation}
P=\min\!\left\{N,\left\lfloor P_0+
\min\!\left(\frac{T}{\tau_p},1\right)P_1\right\rfloor\right\}.
\label{eq:probes}
\end{equation}
We use $P_0$=$P_1$=$256$ and $\tau_p$=$1800$ seconds. This schedule increases temporal resolution for longer videos while bounding the index size. Let $\mathcal{P}$=$\{p_1,\ldots,p_P\}$ be the temporally ordered original-frame indices of the probes. Consecutive groups of $L$ probe intervals form segments $\{\mathcal{G}_1,\ldots,\mathcal{G}_J\}$; we set $L$=$4$. These segments provide coarse temporal units for later query-conditioned refinement.

\noindent\textbf{Low-Resolution Appearance Encoding.}
We resize each probe to a short side of $d=32$ pixels while preserving its aspect ratio, convert it to grayscale, flatten it, and apply $\ell_2$ normalization to obtain $\mathbf{f}_{p_i}$. These cached descriptors support inexpensive appearance comparisons (Section~\ref{sec:appendix-appearance}).

\noindent\textbf{Local Visual-Change Estimation.}
Let $\operatorname{Adj}_{\mathcal{P}}(n)$ contain the available probes immediately before and after frame index $n$ on the sparse grid. We define local visual-change as
\begin{equation}
\Delta_n=\frac{1}{2}
\max_{m\in\operatorname{Adj}_{\mathcal{P}}(n)}
\|\mathbf{f}_n-\mathbf{f}_m\|_2.
\label{eq:visual-change}
\end{equation}
This lightweight, query-independent score highlights probes that differ from their temporal neighbors, directing evidence discovery toward visually distinctive moments.

\noindent\textbf{Visual Observability Assessment.}
We associate each frame with sharpness $s_n^{\mathrm{sharp}}\in[0,1]$ and well-exposedness $\ell_n^{\mathrm{light}}\in[0,1]$. Following \citet{crete2007blur}, sharpness measures how strongly directional local variation is attenuated by controlled reblurring; following \citet{mertens2007exposure}, well-exposedness downweights pixels whose intensities are far from the mid-range, including severely dark or saturated regions. Section~\ref{sec:appendix-observability} gives their detailed definitions. We define the visual observability prior as $O_n$=$s_n^{\mathrm{sharp}}\ell_n^{\mathrm{light}}$. The product requires both adequate sharpness and exposure, preventing one component from compensating for failure of the other. 

\subsection{Query-Conditioned Evidence Discovery}

The index summarizes the video, while the question guides the search for answer-supporting observations. \textsc{MEC} scores the probes and refines only the most promising regions.

\noindent\textbf{Multi-Cue Evidence Scoring.}
Let $\rho_n(q)\in[0,1]$ be the frame--question matching probability produced by the BLIP-2 ITM head \citep{li2023blip2}. We define evidence as the suitability of a frame to serve as an effective observation for $q$ within a compact visual context:
\begin{equation}
E_n(q)=\lambda_r\rho_n(q)
+\lambda_{\Delta}\Delta_n
+\lambda_o O_n,
\label{eq:evidence}
\end{equation}
where $\lambda_r$=$0.7$, $\lambda_{\Delta}$=$0.2$, and $\lambda_o$=$0.1$. Relevance aligns frames with the question, visual change favors distinctive local appearances, and observability downweights visually unclear frames.

\noindent\textbf{Evidence-Guided Segment Activation.}
For segment $\mathcal{G}_j$, let $\bar{E}_j$ be its mean probe evidence score. Given an activation ratio $\eta\in(0,1]$, we retain the $\lceil\eta J\rceil$ segments with the largest $\bar{E}_j$, forming the active set $\mathcal{A}(q)$. We use $\eta$=$0.25$. Aggregating scores by segment reduces sensitivity to isolated high scores and limits fine-grained search to a small part of the timeline.

\noindent\textbf{Anchor-Centered Local Zooming.}
In each active segment $\mathcal{G}_j\in\mathcal{A}(q)$, the highest-evidence probe serves as the anchor $a_j$. Its available preceding and following probes, $a_j^{-}$ and $a_j^{+}$, delimit the two local intervals to be refined. The operator $\operatorname{Uni}(a,b,L_z)$ samples $L_z$ distinct, uniformly spaced interior frame indices between $a$ and $b$, producing the anchor-centered zoom set
\begin{equation}
\mathcal{Z}_j(q)
=\operatorname{Uni}(a_j^{-},a_j,L_z)
\cup\operatorname{Uni}(a_j,a_j^{+},L_z).
\label{eq:zoom}
\end{equation}
We set $L_z$=$2$. Refining both sides of the anchor introduces fine-grained zoom probes that may capture informative observations missed by the global sparse grid.

\noindent\textbf{Candidate Pool Construction.}
The candidate pool $\mathcal{C}(q)$ combines sparse probes $\mathcal{P}$ and query-conditioned zoom probes $\mathcal{Z}(q)$:
$
\mathcal{C}(q)=\mathcal{P}\cup\mathcal{Z}(q).
$
For each zoom probe, $\mathbf{f}_n$, $\Delta_n$, $O_n$, $\rho_n(q)$, and $E_n(q)$ are computed as above. This budget-independent pool supports ranking lengths $M\leq|\mathcal{C}(q)|$.

\subsection{Evidence-to-Context Matryoshka Ranking}

We approximate the marginal decision implied by the all-prefix objective, rather than downstream utility itself. The surrogate combines query-conditioned evidence with temporal and visual information absent from the current prefix. Section~3.1 motivates their relative weights.

\noindent\textbf{Temporal Coverage and Visual Diversity Assessment.}
For rank $k\geq2$, the selected prefix is $\mathcal{S}_{k-1}$=$\{r_1,\ldots,r_{k-1}\}$. An unselected candidate $n$ is assessed by
\begin{align}
H_n(k)=\min\!\left(
1,
\frac{\min_{m\in\mathcal{S}_{k-1}}|n-m|}{N/k}
\right),
\label{eq:coverage}\\
D_n(k)=1-
\max_{m\in\mathcal{S}_{k-1}}\mathbf{f}_n^{\top}\mathbf{f}_m.
\label{eq:diversity}
\end{align}
The interval $N/k$ normalizes distance to the nearest selected frame. The marginal context cues $H_n(k)$ and $D_n(k)$ favor temporal regions and appearances absent from $\mathcal{S}_{k-1}$, respectively.

\noindent\textbf{Prefix-Exposure Weight Scheduling.}
The discrete exposure $\alpha_k$ is a non-increasing step profile determined by $\mathcal{B}$ and $\{\pi_K\}$. We use $1-(k/M)^\gamma$ as a smooth profile shared across budget sets. In a continuous relaxation, this is the survival function of the normalized-budget CDF $F(x)=x^\gamma$ on $[0,1]$; $\gamma=1$ gives a uniform budget prior and linear decay. The evidence weight follows this prior:
\begin{equation}
\begin{aligned}
w_e(k)=w_e^{-}+(w_e^{+}-w_e^{-})
\left[1-\left(\frac{k}{M}\right)^{\gamma}\right],
\end{aligned}
\label{eq:weights}
\end{equation}
Here, $w_e^{+}$ and $w_e^{-}$ specify the upper and lower evidence-weight references, and $\gamma>0$ controls the transition rate. The positive floor $w_e^{-}$ retains evidence sensitivity at large budgets. Temporal coverage receives the remaining adaptive weight, $w_c(k)=1-w_e(k)-w_d$, while $w_d$ stays fixed to control visual redundancy throughout the ranking. We use $w_e^{+}$=$0.6$, $w_e^{-}$=$0.2$, $\gamma$=$1$, and $w_d$=$0.2$. Early ranks emphasize evidence; later ranks favor complementary temporal context.

\begin{table}[t]
\caption{Comparison of frame selectors using Qwen3.5-9B. All values are percentages. Video-MME-v2 reports the nonlinear group score (Non-Lin.) and micro accuracy (Acc.); LVB abbreviates LongVideoBench, and MLVU reports macro accuracy. Best results within each budget are bold.}
\label{tab:main}
\centering
\small
\begingroup
\setlength{\tabcolsep}{3.8pt}
\renewcommand{\arraystretch}{1.04}
\begin{tabularx}{\textwidth}{@{}c l *{8}{>{\centering\arraybackslash}X}@{}}
\toprule
\multirow{2}{*}{$K$} & \multirow{2}{*}{Selector} & \multicolumn{4}{c}{Video-MME} & \multicolumn{2}{c}{Video-MME-v2} & \multirow{2}{*}{LVB} & \multirow{2}{*}{\shortstack{MLVU}} \\
\cmidrule(lr){3-6}\cmidrule(lr){7-8}
& & Short & Medium & Long & Overall & Non-Lin. & Acc. & & \\
\midrule
8 & Uniform & 69.00 & 57.44 & 50.44 & 58.96 & 8.59 & 23.84 & 54.15 & 57.40 \\
8 & AKS & 70.89 & 62.00 & 53.33 & 62.07 & 10.59 & \textbf{26.19} & 61.18 & 66.40 \\
8 & WFS-SB & 73.22 & 64.11 & 53.22 & 63.52 & \textbf{10.93} & 26.06 & 62.00 & 68.33 \\
\oursrow 8 & \textbf{\textsc{MEC}}~\textit{(ours)} & \textbf{73.78} & \textbf{64.33} & \textbf{53.44} & \textbf{63.85} & 10.81 & 26.13 & \textbf{62.45} & \textbf{69.46} \\
\midrule
16 & Uniform & 73.67 & 60.67 & 53.56 & 62.63 & 10.06 & 25.44 & 57.52 & 60.29 \\
16 & AKS & 75.67 & 65.00 & 54.56 & 65.07 & 11.30 & 27.03 & 60.81 & 68.65 \\
16 & WFS-SB & 76.56 & 65.56 & 55.22 & 65.78 & \textbf{12.23} & \textbf{28.22} & 61.93 & \textbf{68.94} \\
\oursrow 16 & \textbf{\textsc{MEC}}~\textit{(ours)} & \textbf{77.22} & \textbf{66.89} & \textbf{55.44} & \textbf{66.52} & 12.07 & 27.78 & \textbf{62.30} & 68.39 \\
\midrule
32 & Uniform & 77.78 & 65.11 & 55.56 & 66.15 & 11.41 & 27.47 & 58.86 & 65.00 \\
32 & AKS & 79.00 & 68.67 & 57.00 & 68.22 & 12.95 & 29.50 & 61.33 & 70.43 \\
32 & WFS-SB & \textbf{79.56} & 68.56 & 57.56 & 68.56 & \textbf{13.42} & \textbf{29.69} & 63.35 & \textbf{70.90} \\
\oursrow 32 & \textbf{\textsc{MEC}}~\textit{(ours)} & 78.78 & \textbf{70.33} & \textbf{58.22} & \textbf{69.11} & 13.05 & 29.47 & \textbf{63.58} & 70.36 \\
\midrule
64 & Uniform & 80.44 & 68.78 & 59.22 & 69.48 & 12.79 & 28.78 & 60.73 & 69.69 \\
64 & AKS & 81.33 & 72.33 & 59.78 & 71.15 & 14.23 & 31.00 & 64.62 & 72.46 \\
64 & WFS-SB & 80.67 & 71.44 & \textbf{61.67} & 71.26 & 14.64 & 31.31 & \textbf{66.57} & 73.83 \\
\oursrow 64 & \textbf{\textsc{MEC}}~\textit{(ours)} & \textbf{81.89} & \textbf{73.00} & 60.56 & \textbf{71.81} & \textbf{14.77} & \textbf{31.69} & 65.89 & \textbf{74.30} \\
\midrule
128 & Uniform & \textbf{83.00} & 73.11 & 61.33 & 72.48 & 14.33 & 31.00 & 65.00 & 73.77 \\
128 & AKS & 82.56 & 74.67 & \textbf{62.78} & 73.33 & 15.50 & 32.47 & 67.61 & 74.88 \\
128 & WFS-SB & 82.56 & 75.11 & 62.44 & 73.37 & \textbf{15.71} & \textbf{33.16} & 68.06 & \textbf{75.85} \\
\oursrow 128 & \textbf{\textsc{MEC}}~\textit{(ours)} & 82.56 & \textbf{75.89} & \textbf{62.78} & \textbf{73.74} & 15.42 & 33.00 & \textbf{68.36} & \textbf{75.85} \\
\midrule
256 & Uniform & 82.33 & 74.33 & 64.44 & 73.70 & 15.78 & 32.75 & 66.87 & 76.01 \\
256 & AKS & 82.67 & 75.89 & 64.44 & 74.33 & 16.31 & 33.75 & 68.21 & 76.19 \\
256 & WFS-SB & 82.67 & \textbf{76.11} & 64.00 & 74.26 & 16.81 & 34.56 & \textbf{69.78} & \textbf{77.01} \\
\oursrow 256 & \textbf{\textsc{MEC}}~\textit{(ours)} & \textbf{83.11} & 75.56 & \textbf{65.22} & \textbf{74.63} & \textbf{17.08} & \textbf{34.63} & 68.51 & 76.65 \\
\bottomrule
\end{tabularx}
\endgroup

\vspace{-10pt}
\end{table}

\noindent\textbf{Greedy Matryoshka Ranking.}
At rank $k\geq2$, we score each remaining candidate $n$ as
\begin{equation}
u_k(n)=w_e(k)E_n(q)+w_c(k)H_n(k)+w_dD_n(k).
\label{eq:greedy-ranking}
\end{equation}
We interpret $u_k(n)$ as a tractable proxy for the unobservable marginal contribution of assigning $n$ to rank $k$ under Equation~\eqref{eq:matryoshka-objective}, conditioned on $\mathcal{S}_{k-1}$. Evidence $E_n(q)$ captures answer-supporting potential, while $H_n(k)$ and $D_n(k)$ capture complementarity to the current prefix. The score is position-aware through its weights and set-aware through these context terms. Starting from the highest-evidence candidate, the ranking greedily appends the candidate with the largest $u_k(n)$. Selected frames are never replaced, ensuring exact nesting without combinatorial lookahead. Algorithm~\ref{alg:greedy-matryoshka} gives the complete procedure; Section~4.3 evaluates the surrogate structure empirically. For any budget $K\leq M$, \textsc{MEC} chronologically sorts $\mathcal{R}_{:K}=(r_1,\ldots,r_K)$ before LMM inference.

\begin{figure}[t]
\centering
\includegraphics[width=\textwidth]{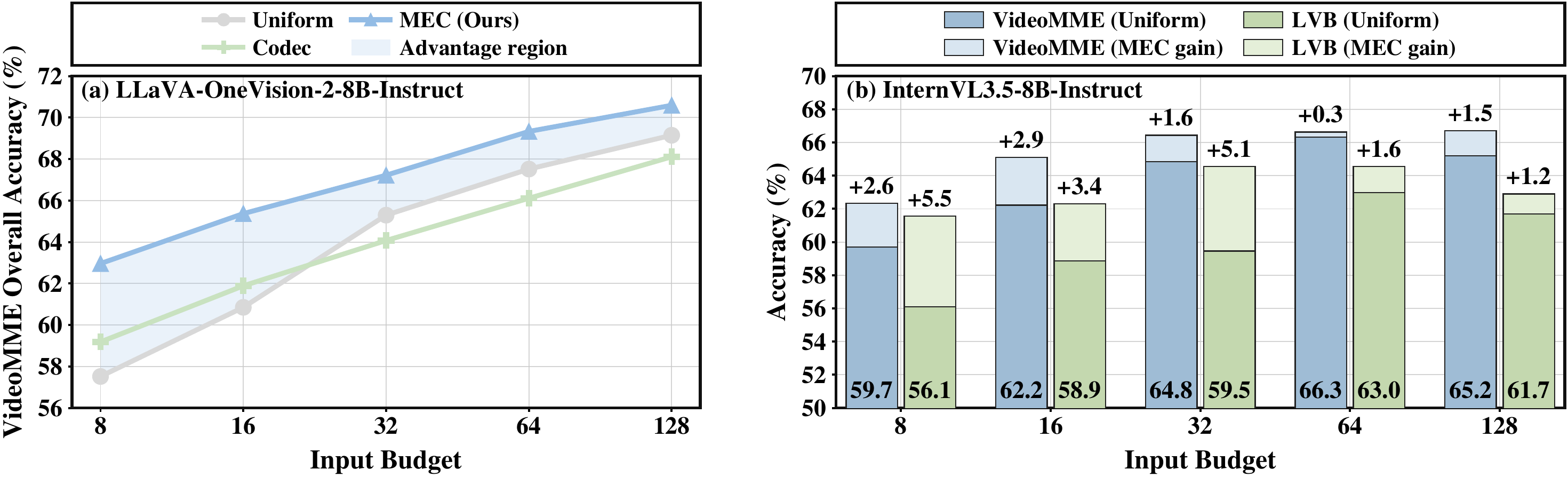}
\caption{Generalization across downstream LMMs and input budgets $K\in\{8,16,32,64,128\}$. (a) Video-MME Overall with LLaVA-OneVision-2-8B-Instruct, comparing uniform frames, native codec-stream input, and \textsc{MEC}. (b) Uniform accuracy and the corresponding \textsc{MEC} gain on Video-MME and LongVideoBench (LVB) with InternVL3.5-8B-Instruct.}
\label{fig:cross-model}

\medskip
\centering
\includegraphics[width=0.99\textwidth]{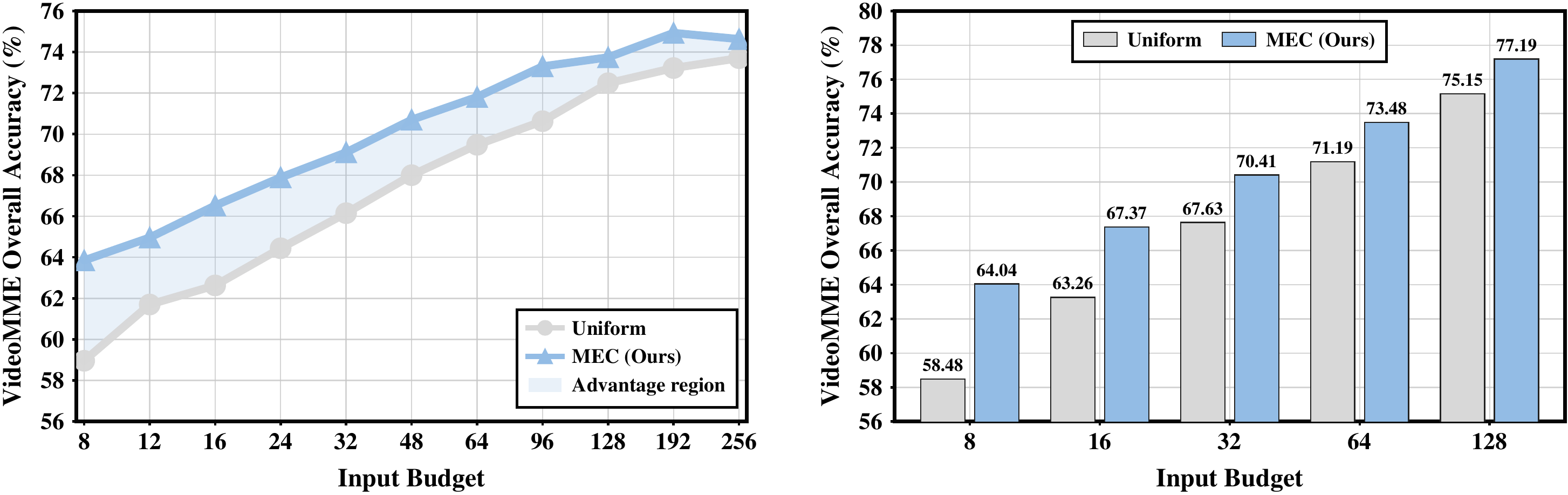}
\caption{Budget behavior and thinking-mode evaluation on Video-MME Overall with Qwen3.5-9B. Left: Uniform and \textsc{MEC} at 11 budgets; shading marks the \textsc{MEC} gain. Right: the two selectors at five budgets with thinking enabled.}
\label{fig:fine-grained-budget}
\vspace{-10pt}
\end{figure}
\vspace{-5pt}

\section{Experiments}
\vspace{-5pt}

\subsection{Experimental Settings}
\vspace{-5pt}

\noindent\textbf{Benchmarks.}
We evaluate on four complementary long-video benchmarks. Video-MME contains 900 videos and 2,700 multiple-choice questions across short-, medium-, and long-duration groups \citep{fu2025videomme}. Video-MME-v2 contains 800 videos and 3,200 questions organized around information aggregation, temporal dynamics, and complex reasoning; in addition to micro accuracy, its nonlinear group score measures consistency across related questions \citep{fu2026videommev2}. LongVideoBench contains 3,763 videos and 6,678 questions that emphasize long-context referring and temporal reasoning \citep{wu2024longvideobench}. MLVU contains 1,730 videos, 3,102 question--answer pairs, and nine tasks spanning global and local understanding \citep{zhou2025mlvu}. Section~\ref{sec:appendix-benchmarks} provides additional benchmark scope and task details.

\noindent\textbf{Implementation Details.}
We evaluate with LMMs-Eval \citep{zhang2025lmmseval} on 8 NVIDIA H20 GPUs. The downstream LMMs are Qwen3.5-9B, LLaVA-OneVision-2-8B-Instruct, and InternVL3.5-8B-Instruct. We compare with uniform sampling, AKS \citep{tang2025aks}, and WFS-SB \citep{chen2026wfssb}. All methods use the same question text, downstream prompt and dataset-specific scoring protocol. Every \textsc{MEC} experiment uses one shared configuration across all benchmarks, downstream LMMs, and frame budgets, without model-, dataset-, or budget-specific tuning. We set $M$=$256$, $(\lambda_r,\lambda_\Delta,\lambda_o)$=$(0.7,0.2,0.1)$, and $(w_e^{+},w_e^{-},w_d,\gamma)$=$(0.6,0.2,0.2,1)$. Baselines follow their released procedures and default settings. Table~\ref{tab:appendix-configuration} gives the full configuration.

\vspace{-5pt}
\subsection{Main Results}
\vspace{-5pt}

\noindent\textbf{Comparison with State-of-the-Art.}
We first compare \textsc{MEC} with training-free frame selectors using Qwen3.5-9B over six budgets. Table~\ref{tab:main} shows that the benefit of \textsc{MEC} extends across datasets and budgets. Averaging Video-MME Overall, Video-MME-v2 micro accuracy, LongVideoBench accuracy, and MLVU accuracy over all six budgets, \textsc{MEC} improves uniform sampling by $3.77$ pp. The advantage is especially pronounced for tight prefixes: at $K$=$8$, the gains over uniform sampling are $8.30$ pp on LongVideoBench and $12.06$ pp on MLVU. Against stronger selectors, \textsc{MEC} obtains the best Video-MME Overall result at every budget, including $66.52$ versus $65.78$ at $K$=$16$. WFS-SB remains strongest in several Video-MME-v2, LongVideoBench, and MLVU settings, but \textsc{MEC} stays competitive while uniquely serving all budgets with one shared ranking.

\noindent\textbf{Generalization Across LMMs.}
We next test whether the same selector configuration transfers across architectures and input representations.
With LLaVA-OneVision-2, \textsc{MEC} outperforms uniform frames and codec-stream input at every budget. Codec leads uniform sampling at $K$=$8$ and $16$, but falls behind from $K$=$32$. Codec-stream tokenization follows bit-cost dynamics and motion-residual cues to concentrate capacity on event-bearing intervals \citep{an2026llavaov2}. We interpret the crossover as a possible consequence of this allocation: it may favor sparse motion-relevant evidence at small budgets, while additional decoded frames may preserve more local texture, fine-grained static evidence, and continuous trajectories. \textsc{MEC} remains strongest across the crossover, showing that query-conditioned evidence discovery and evidence-to-context organization improve over model-specific input allocation while retaining original-frame evidence.
The InternVL3.5 results in Figure~\ref{fig:cross-model}(b) provide a complementary test. \textsc{MEC} improves over uniform sampling at all five budgets on both Video-MME and LongVideoBench, despite using exactly the same configuration as the other experiments. The gains transfer across downstream architecture, benchmark, and context size rather than being tied to Qwen3.5-9B or a single evaluation set.

\begin{table}[t]
\caption{Component and ranking-schedule ablations on Video-MME Overall with Qwen3.5-9B. Low, Med., and High average $K=\{8,16\}$, $\{32,64\}$, and $\{128,256\}$; Avg. covers all six budgets. In (b), fixed-weight tuples give $(w_e,w_c,w_d)$; decay schedules use $(w_e^{+},w_e^{-})=(.6,.2)$, $w_d=.2$ and $w_c(k)=1-w_e(k)-w_d$. Best values in each panel are bold.}
\label{tab:ablation}
\centering
\begin{subtable}[t]{0.49\linewidth}
\caption{Component contributions}
\label{tab:ablation-components}
\centering
\small
\setlength{\tabcolsep}{2.3pt}
\renewcommand{\arraystretch}{1.06}
\begin{tabularx}{\linewidth}{@{}>{\raggedright\arraybackslash}Xrrrr@{}}
\toprule
Variant & Low & Med. & High & Avg. \\
\midrule
Uniform & 60.80 & 67.81 & 73.09 & 67.23 \\
\oursrow \textbf{\textsc{MEC}} & \textbf{65.19} & \textbf{70.46} & \textbf{74.19} & \textbf{69.94} \\
w/o coverage & 61.69 & 68.13 & 73.63 & 67.81 \\
w/o diversity & 64.63 & 70.06 & 73.72 & 69.47 \\
w/o local zoom & 64.26 & 69.94 & 73.91 & 69.37 \\
w/o visual change & 64.11 & 69.59 & 73.85 & 69.19 \\
w/o observability & 64.44 & 70.26 & 74.04 & 69.58 \\
\bottomrule
\end{tabularx}
\end{subtable}\hfill
\begin{subtable}[t]{0.49\linewidth}
\caption{Ranking-weight schedules}
\label{tab:ablation-schedules}
\centering
\small
\setlength{\tabcolsep}{2.3pt}
\renewcommand{\arraystretch}{1.06}
\begin{tabularx}{\linewidth}{@{}>{\raggedright\arraybackslash}Xrrrr@{}}
\toprule
Schedule & Low & Med. & High & Avg. \\
\midrule
Uniform & 60.80 & 67.81 & 73.09 & 67.23 \\
\oursrow \textbf{\textsc{MEC}} ($\gamma=1$) & \textbf{65.19} & 70.46 & 74.19 & \textbf{69.94} \\
Fixed weights $(.6,.2,.2)$ & 64.89 & 70.09 & 73.56 & 69.51 \\
Fixed weights $(.2,.6,.2)$ & 63.44 & 69.46 & 74.06 & 68.99 \\
Fixed weights $(.4,.4,.2)$ & 64.66 & 70.15 & 73.82 & 69.54 \\
Decay $\gamma=1.5$ & 64.70 & \textbf{70.61} & \textbf{74.22} & 69.85 \\
Decay $\gamma=2/3$ & 64.49 & 70.22 & 73.89 & 69.60 \\
\bottomrule
\end{tabularx}
\end{subtable}

\medskip
\centering
\begin{minipage}[t]{0.54\linewidth}
\caption{Ranking horizon $M$ vs. budget $K$ on Video-MME Overall. Bold: column best; --: $K>M$.}
\label{tab:ranking-length}
\centering
\small
\setlength{\tabcolsep}{2pt}
\renewcommand{\arraystretch}{1.06}
\begin{tabularx}{\linewidth}{@{}c *{6}{>{\raggedleft\arraybackslash}X}@{}}
\toprule
$M\backslash K$ & 8 & 16 & 32 & 64 & 128 & 256 \\
\midrule
8   & 62.44 & -- & -- & -- & -- & -- \\
16  & 63.48 & \textbf{66.70} & -- & -- & -- & -- \\
32  & \textbf{64.04} & 65.78 & 68.26 & -- & -- & -- \\
64  & 63.89 & 66.11 & \textbf{69.22} & 71.04 & -- & -- \\
128 & 63.52 & 66.00 & 68.52 & 71.70 & 73.26 & -- \\
\oursrow \textbf{256} & 63.85 & 66.52 & 69.11 & \textbf{71.81} & \textbf{73.74} & \textbf{74.63} \\
\bottomrule
\end{tabularx}
\end{minipage}\hfill
\begin{minipage}[t]{0.43\linewidth}
\caption{Mean time costs over six budgets on Video-MME Long.}
\label{tab:efficiency}
\centering
\small
\setlength{\tabcolsep}{2pt}
\renewcommand{\arraystretch}{1.06}
\begin{tabularx}{\linewidth}{@{}l *{3}{>{\raggedleft\arraybackslash}X}@{}}
\toprule
 & \textbf{\textsc{MEC}} & AKS & WFS-SB \\
\midrule
Candidates & \textbf{635.40} & 2,466.82 & 2,466.82 \\
Prep. (s) & \textbf{5.48} & 12.09 & 12.09 \\
Sel. (s) & \textbf{7.89} & 22.06 & 25.50 \\
E2E (s) & \textbf{23.09} & 43.87 & 47.31 \\
\bottomrule
\end{tabularx}
\par\smallskip
\raggedright Prep.: preprocessing; Sel.: selection; E2E: end-to-end time per query.
\end{minipage}
\vspace{-5pt}
\end{table}

\vspace{-5pt}
\subsection{Further Quantitative Analyses}

\noindent\textbf{Fine-Grained Budget Behavior.}
We evaluate eleven budgets to determine whether the gains persist between the six operating points used in Table~\ref{tab:main}.
Figure~\ref{fig:fine-grained-budget} shows a positive gain at every budget from $K$=$8$ to $K$=$256$, including five intermediate values absent from Table~\ref{tab:main}. The margin is largest at $K$=$8$ ($+4.89$ pp) and remains positive at $K$=$256$ ($+0.93$ pp). Although the gap is not strictly monotonic, it generally narrows as the input grows: tight prefixes benefit most from concentrating diagnostic evidence, whereas larger inputs allow uniform sampling to recover more context. The persistent residual gain indicates that the later ranks of \textsc{MEC} continue to organize context more effectively rather than merely exploiting a few favorable budget choices.

\noindent\textbf{Thinking-Mode Evaluation.}
For Figure~\ref{fig:fine-grained-budget} (right), we enable Qwen3.5 thinking mode through its official interface \citep{qwen2026qwen35}; standard evaluations disable it. \textsc{MEC} improves over Uniform at all five tested budgets, with gains decreasing from $5.56$ pp at $K=8$ to $2.04$ pp at $K=128$ (Section~\ref{sec:appendix-thinking}). Frame selection remains useful with downstream reasoning enabled.

\noindent\textbf{Component and Ranking Ablations.}
Table~\ref{tab:ablation} shows the complementary contributions of candidate discovery and prefix construction. Removing coverage, diversity, local zoom, visual change $\Delta_n$, or observability $O_n$ reduces average accuracy by $2.13$, $0.47$, $0.57$, $0.75$, and $0.36$ pp, respectively. The full configuration improves Low, Medium, and High accuracy over Uniform by $4.39$, $2.65$, and $1.10$ pp. With candidate discovery and downstream evaluation fixed, the position-adaptive ranking exceeds all three fixed-weight configurations by $0.40$--$0.95$ pp on average, supporting different surrogate weights across positions under Equation~\eqref{eq:matryoshka-objective}. Table~\ref{tab:appendix-ranking-ablation} further shows that reversing the evidence-to-context transition reduces average accuracy from $69.94$ to $69.00$, whereas the five tested evidence-decay schedules span only $0.46$ pp across decay rates and endpoints. These results support the evidence-first, context-later structure more strongly than any particular decay curve. Together with the cross-model and cross-benchmark results in Figure~\ref{fig:cross-model} and Table~\ref{tab:main}, they support using one default configuration across models, benchmarks, and budgets.

\begin{figure}[t]
\centering
\setlength{\qualitativeimageheight}{0.40\linewidth}
\begin{subfigure}[t]{0.49\linewidth}
\centering
\begin{minipage}[c][\qualitativeimageheight][c]{\linewidth}
\includegraphics[width=\linewidth]{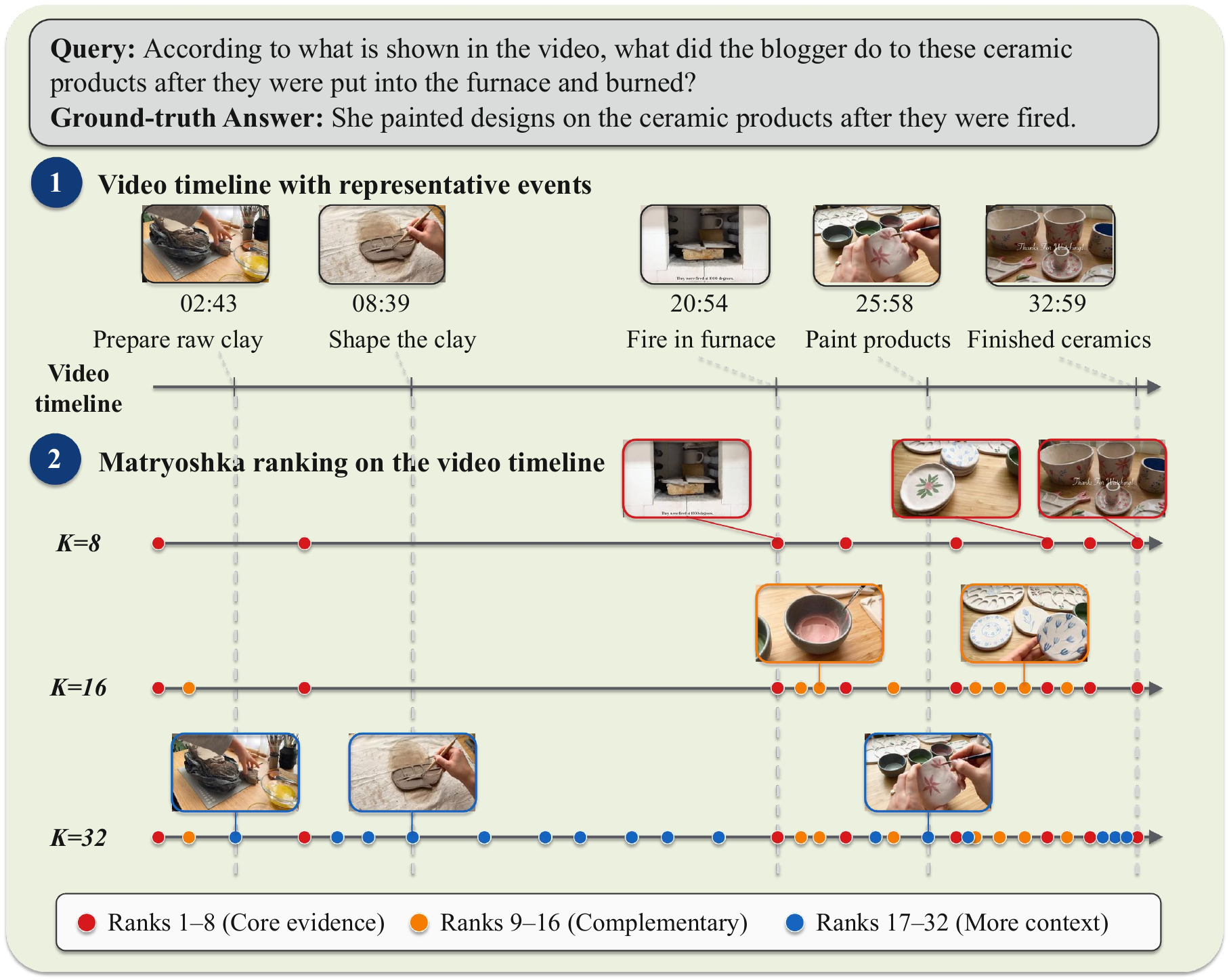}
\end{minipage}
\caption{Evidence-to-context expansion}
\label{fig:visualization}
\end{subfigure}\hfill
\begin{subfigure}[t]{0.49\linewidth}
\centering
\begin{minipage}[c][\qualitativeimageheight][c]{\linewidth}
\includegraphics[width=\linewidth]{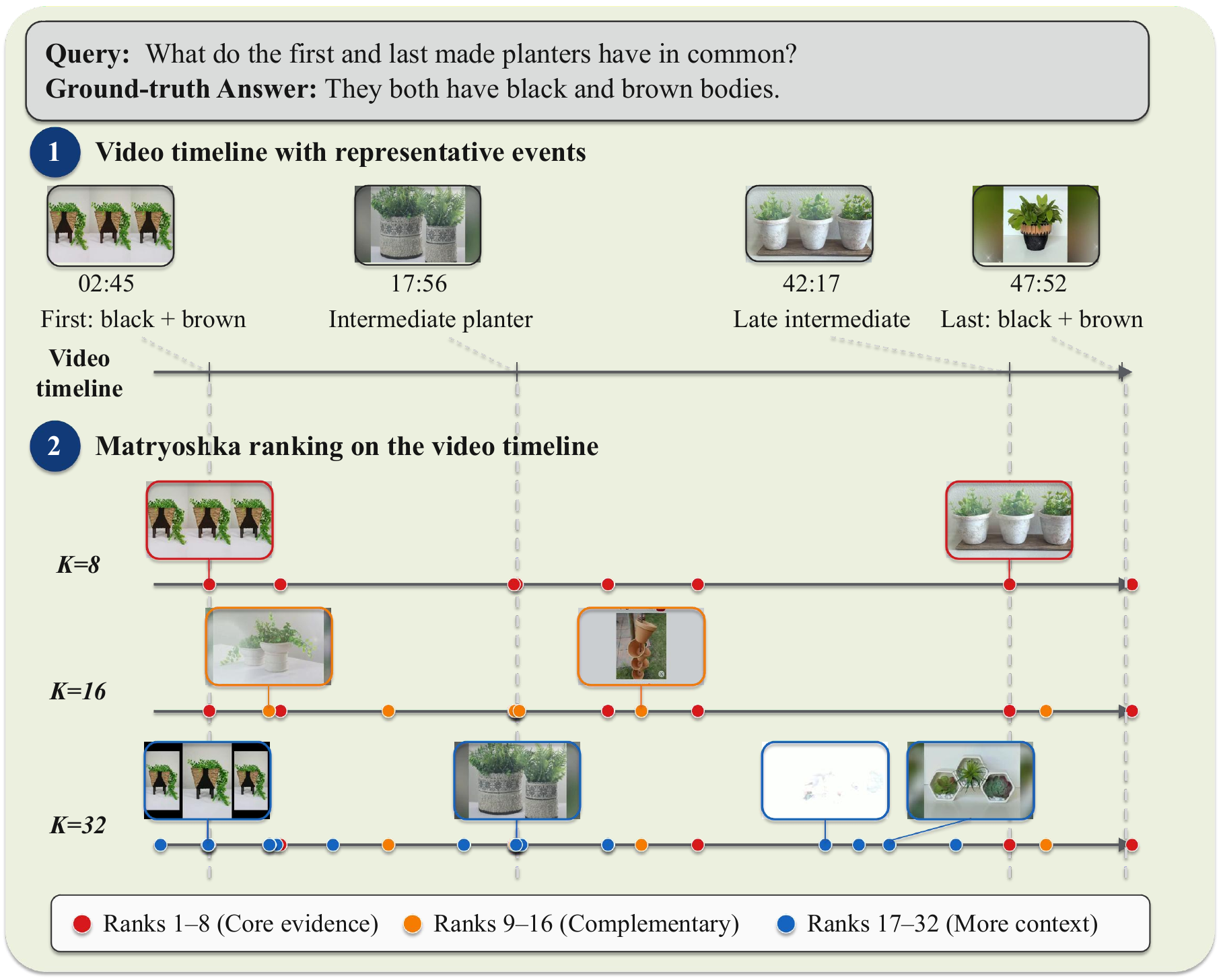}
\end{minipage}
\caption{Missed endpoint evidence}
\label{fig:failure-case}
\end{subfigure}
\caption{Success and failure of \textsc{MEC}. (a) Larger prefixes retain ceramic-making evidence and add earlier preparation stages. (b) Sparse discovery misses a clear view of the last completed planter, leaving the first--last comparison unresolved. Red, orange, and blue mark frames introduced at ranks 1--8, 9--16, and 17--32. All prefixes share one ranking and are shown chronologically.}
\label{fig:qualitative}
\vspace{-10pt}
\end{figure}
\vspace{-3pt}

\noindent\textbf{Effect of Ranking Horizon.}
Table~\ref{tab:ranking-length} compares a shared long ranking with rankings terminated at each target budget. Setting $M=K$ completes the evidence-to-context transition within that budget; $M=256$ keeps short prefixes focused on evidence and reserves later positions for context expansion. Across $K\in\{8,16,32,64,128\}$, the shared ranking improves mean accuracy by $0.67$ pp over the $M=K$ diagonal, including gains of $1.41$, $0.85$, $0.77$, and $0.48$ pp at $K=8,32,64,$ and $128$. It achieves the best results at $K=64$ and $128$ and stays within $0.19$ pp of the best at each smaller budget. A single Matryoshka ranking therefore provides an accuracy benefit across budgets while eliminating the need to construct a new sequence for each one.

\noindent\textbf{Selector Efficiency.}
Table~\ref{tab:efficiency} reports time costs on Video-MME Long with cached RGB frames that can be reused across questions about the same video. For a fair comparison at each budget, the reported Sel. cost includes an independent run of \textsc{MEC} selection for every evaluated budget. In deployment, changing the budget for the same query incurs no additional selector cost: the existing ranking is truncated to the required prefix and the selected frames are restored to chronological order. \textsc{MEC} evaluates $635.40$ candidates on average, $74.24\%$ fewer than AKS and WFS-SB, and reduces preprocessing from $12.09$ to $5.48$ seconds per query. Selection takes $7.89$ seconds, compared with $22.06$ for AKS and $25.50$ for WFS-SB, reductions of $64.23\%$ and $69.06\%$. Including $9.72$ seconds of LMM inference, end-to-end latency falls from $43.87$ and $47.31$ seconds to $23.09$ seconds, respectively. These savings accompany the accuracy gains in Figure~\ref{fig:overview}.

\vspace{-3pt}
\subsection{Qualitative Analysis}
\vspace{-3pt}

In Figure~\ref{fig:visualization}, the $K=8$ prefix concentrates on the furnace, painted surfaces, and finished ceramics, providing the evidence that painting follows firing. The $K=16$ prefix retains these frames and adds pigment and complementary product views; $K=32$ further covers raw-clay preparation and shaping. Larger budgets reveal more of the manufacturing process while preserving answer-relevant observations. Figure~\ref{fig:failure-case} shows a first--last comparison in a $48.3$-minute video. The first completed planter at $164.6$ s and the last at $2871.7$ s both have black-and-brown bodies. Sparse probes capture the first planter, intermediate products, and observations near the end, but miss a clear view of the last completed planter. This brief state receives no anchor for local zooming and never enters the candidate pool. The LMM answers incorrectly with \textsc{MEC} at all six budgets, while Uniform yields correct answers from $K=32$ onward. All prefixes draw from the same pool, so later ranks add coverage and diversity only among discovered frames. Candidate recall, particularly for brief endpoint events, is the bottleneck: larger prefixes cannot recover evidence missed during discovery.

\section{Conclusion}

We formulate long-video frame selection as a Matryoshka ranking problem: one query-conditioned sequence serves multiple budgets, concentrating evidence in short prefixes and expanding context in longer ones. We introduce \textsc{MEC}, a training-free framework that combines reusable sparse indexing, query-conditioned evidence discovery, and position-adaptive ranking to construct this sequence efficiently. Experiments across four benchmarks, three downstream LMMs, and six frame budgets validate its effectiveness, generalization, and favorable accuracy--latency trade-off. By decoupling frame selection from the target budget, \textsc{MEC} offers a practical foundation for long-video systems that must adapt their visual context to changing deployment constraints.

\subsection*{AI use statement}
In this work, we used generative AI tools to polish the manuscript for clarity and readability and to assist in searching for and identifying relevant literature. We take responsibility for the final content of this work, including all AI-assisted text and citations.

\bibliographystyle{plainnat}
\bibliography{references}

\clearpage
\appendix

\noindent\textbf{Appendix overview.}
Sections~\ref{sec:appendix-related-work} and~\ref{sec:appendix-experimental-details} provide additional related work and experimental details. Section~\ref{sec:appendix-method-details} specifies the appearance descriptor, observability measures, and ranking algorithm. Section~\ref{sec:appendix-limitations} discusses limitations, and Section~\ref{sec:appendix-additional-experiments} presents further ablations, model-scale and thinking-mode results, and a qualitative example.

\section{Additional Related Work}
\label{sec:appendix-related-work}

\noindent\textbf{Video Large Multimodal Models.}
Video LMMs encode frames or clips as visual tokens for joint visual, temporal, and language reasoning. Recent systems improve this interface through dynamic visual resolution, spatiotemporal position modeling, grounded video supervision, tracking, and codec-aware tokenization \citep{bai2025qwen3vl,wang2025internvl35,clark2026molmo2,an2026llavaov2}. These advances do not eliminate the need for input selection: redundant observations can still occupy the input while brief or subtle events are missed. \textsc{MEC} selects original frames and treats the downstream LMM as a black box. The selector can therefore be used across architectures without access to internal token scores.

\noindent\textbf{Matryoshka Representations.}
Matryoshka Representation Learning arranges an embedding so that nested dimensional prefixes retain progressively finer information; MatFormer extends this idea to nested Transformer subnetworks \citep{kusupati2022matryoshka,devvrit2024matformer}. For multimodal inputs, Matryoshka Multimodal Models learn nested visual-token sets, and the Matryoshka Query Transformer orders latent queries whose prefixes yield different token budgets \citep{cai2025matryoshka,hu2024mqt}. \textsc{MEC} applies the nesting principle to a query-conditioned ordering of original video-frame indices constructed at inference time. Early ranks concentrate answer-relevant evidence, while later ranks add temporal and visual context.

\noindent\textbf{Visual-Token Compression.}
Visual-token compression reduces the input length through temporal aggregation, token merging or pruning, and adaptive budget allocation. Examples include hierarchical clip-to-video compression, joint similarity and importance scoring, diversity-based pruning, and query-oriented token assignment \citep{li2026videochatflash,fu2025framefusion,alvar2025divprune,luo2025quota,zeng2026wavezip}. These methods operate within or around the visual-token pipeline and may use representations unavailable through a frame-only interface. \textsc{MEC} selects a nested sequence of original frames and leaves tokenization to the downstream model. It preserves the visual content of the selected frames until that step; token compression may discard fine details during input reduction.

\section{Additional Experimental Details}
\label{sec:appendix-experimental-details}

\subsection{Benchmarks}
\label{sec:appendix-benchmarks}

\paragraph{Video-MME.}
Video-MME contains 900 videos and 2,700 manually annotated multiple-choice questions, with three questions per video. The videos cover six domains and 30 subfields and range from 11 seconds to one hour \citep{fu2025videomme}. We report accuracy on the official short, medium, and long splits, together with Overall accuracy (Table~\ref{tab:main}). These splits allow us to compare selection performance across video durations.

\paragraph{Video-MME-v2.}
Video-MME-v2 contains 800 videos and 3,200 questions arranged in correlated groups. Its three-level hierarchy progresses from aggregating visual information across multiple points to modeling temporal dynamics and performing complex multimodal reasoning \citep{fu2026videommev2}. We report micro accuracy over individual questions and the official nonlinear group score. The group score rewards consistent correctness across related questions and penalizes partially correct groups, complementing question-level accuracy.

\paragraph{LongVideoBench.}
LongVideoBench contains 3,763 videos and 6,678 human-authored multiple-choice questions in 17 categories. Videos cover movies, news, daily life, and knowledge-oriented content, with four duration ranges spanning roughly ten seconds to one hour \citep{wu2024longvideobench}. The benchmark pairs video with subtitle context and asks models to locate a referenced context before reasoning about its details. We report official question-level accuracy and use video inputs without subtitles, as specified in Table~\ref{tab:appendix-configuration}.

\paragraph{MLVU.}
MLVU contains 1,730 videos and 3,102 question--answer pairs across nine tasks. Videos range from three minutes to two hours, average approximately 15 minutes, and include movies, surveillance footage, egocentric recordings, cartoons, and games \citep{zhou2025mlvu}. The full benchmark covers global understanding and localized reasoning in both multiple-choice and open-ended formats. Under our evaluation protocol, we report macro accuracy over the evaluated tasks, giving each task equal weight regardless of its number of questions.

\subsection{Downstream Large Multimodal Models}
\label{sec:appendix-downstream-models}

\paragraph{Qwen3.5-9B.}
Qwen3.5-9B is a post-trained, 9-billion-parameter model with a unified image, video, and text interface. Its release describes early-fusion multimodal training, a vision encoder coupled to a causal language model, a hybrid Gated DeltaNet and attention backbone, and a native context length of 262,144 tokens \citep{qwen2026qwen35}. Thinking is enabled by default and can be disabled through the official interface. We disable it for the four-benchmark comparison, component and schedule ablations, and efficiency analysis; Figure~\ref{fig:fine-grained-budget} (right) evaluates thinking separately. \textsc{MEC} supplies selected frames without accessing or modifying internal visual tokens.

\paragraph{LLaVA-OneVision-2-8B-Instruct.}
LLaVA-OneVision-2 uses a native OneVision encoder with windowed attention and shared 3D rotary position encoding for images, sampled frames, and codec-derived canvases. Its codec-stream pathway uses bit-cost dynamics to group temporal content and motion-residual cues to allocate spatial evidence within a visual-token budget \citep{an2026llavaov2}. Uniform and \textsc{MEC} use decoded frames. Figure~\ref{fig:cross-model}(a) also evaluates native codec-stream input, allowing us to compare query-conditioned frame selection with the model's compressed-video representation.

\paragraph{InternVL3.5-8B-Instruct.}
InternVL3.5 uses cascade reinforcement learning and a Visual Resolution Router to adjust visual-token resolution. Its release also describes decoupled vision and language deployment across devices \citep{wang2025internvl35}. We use the 8B instruction-tuned checkpoint to test transfer across architectures. Figure~\ref{fig:cross-model}(b) reports Video-MME and LongVideoBench results with the same \textsc{MEC} configuration used elsewhere, without model-specific selector tuning.

\subsection{Comparison Methods}
\label{sec:appendix-comparison-methods}

\paragraph{Adaptive Keyframe Sampling (AKS).}
AKS selects a fixed number of keyframes by balancing query relevance and temporal coverage \citep{tang2025aks}. A lightweight vision--language scorer measures prompt--frame relevance, and recursively partitioned temporal bins measure coverage. The adaptive procedure ranges from selecting the highest-relevance frames to enforcing increasingly uniform coverage. Because selection depends on the target budget, AKS is rerun for each $K$. We use the released procedure and default settings at the same budgets as \textsc{MEC}.

\paragraph{Wavelet-based Frame Selection by Detecting Semantic Boundary (WFS-SB).}
WFS-SB applies multi-resolution wavelet analysis to a temporal query--frame relevance signal, suppressing high-frequency noise and identifying coarse semantic boundaries \citep{chen2026wfssb}. It allocates the target frame budget among the resulting clips according to their importance, then uses maximal marginal relevance to balance relevance and diversity within each clip. Both allocation and within-clip selection depend on the prescribed budget $K$. We use the released procedure and default settings without dataset- or budget-specific retuning.

\subsection{Complete Experimental Configuration}

We run all evaluations through LMMs-Eval \citep{zhang2025lmmseval}. Except where an ablation specifies otherwise, \textsc{MEC} uses the configuration in Table~\ref{tab:appendix-configuration} for every benchmark, downstream model, and frame budget. The candidate pool and length-$M$ ranking are constructed independently of the target inference budget $K$.

\begin{table}[htbp]
\caption{Complete shared configuration for the main \textsc{MEC} experiments.}
\label{tab:appendix-configuration}
\centering
\small
\begingroup
\setlength{\tabcolsep}{3pt}
\renewcommand{\arraystretch}{1.04}
\begin{tabularx}{\linewidth}{@{}lX@{}}
\toprule
Component & Shared setting \\
\midrule
Hardware & 8 NVIDIA H20 GPUs; Intel Xeon Platinum 8575C, 192 cores \\
Video decoding & Decord with \texttt{DECORD\_THREADS=16} \\
Evaluation input & Decoded frames and question text; no subtitles; thinking disabled except for Figure~\ref{fig:fine-grained-budget} (right) \\
Ranking and budgets & $M=256$; $K\in\{8,16,32,64,128,256\}$ \\
Sparse probes & $P_0=P_1=256$; $\tau_p=1800$ s \\
Temporal segments & $L=4$ probe intervals \\
Appearance encoding & Short side $d=32$, grayscale, $\ell_2$ normalization \\
Evidence scorer & BLIP-2 ITM \citep{li2023blip2} \\
Evidence mixture & $(\lambda_r,\lambda_\Delta,\lambda_o)=(0.7,0.2,0.1)$ \\
Segment activation & $\eta=0.25$ \\
Local zoom & $L_z=2$ interior frames on each available side \\
Ranking schedule & $(w_e^{+},w_e^{-},w_d,\gamma)=(0.6,0.2,0.2,1)$ \\
Sharpness & One-dimensional mean reblurring kernel, $L_b=9$ \\
Well-exposedness & $(\mu,\sigma)=(0.5,0.2)$ \\
\bottomrule
\end{tabularx}
\endgroup

\end{table}

For the reported multiple-choice evaluations, the original question and choices are followed by the same post-prompt for every selector and downstream model:
\begin{quote}
\small\ttfamily\raggedright
Answer with the option's letter from the given choices directly.
\end{quote}
This post-prompt specifies only the answer format. Dataset scoring, answer normalization, and other model-side evaluation procedures follow the corresponding LMMs-Eval task implementation. We use no subtitles and disable thinking except in Figure~\ref{fig:fine-grained-budget} (right). No additional model-, dataset-, or budget-specific decoding adjustments are made.

\section{Additional Method Details}
\label{sec:appendix-method-details}

\subsection{Low-Resolution Appearance Encoding}
\label{sec:appendix-appearance}

We encode each probe as
\begin{equation}
\mathbf{f}_{p_i}=
\operatorname{L2-Norm}\!\left(
\operatorname{Vec}\!\left(
\operatorname{Gray}\!\left(\operatorname{Resize}_d(x_{p_i})\right)
\right)\right).
\label{eq:descriptor}
\end{equation}
Here, $\operatorname{Resize}_d$ preserves the aspect ratio and sets the short side to $d=32$ pixels; $\operatorname{Gray}$ and $\operatorname{Vec}$ denote grayscale conversion and vectorization. We cache the normalized descriptors so that appearance comparisons use compact vectors instead of original-resolution frames, reducing storage and computation.

\subsection{Observability Measures}
\label{sec:appendix-observability}

The sparse index estimates observability from query-independent sharpness and exposure measures. Let $\Omega_n$ be the pixel domain after resizing frame $x_n$ to short side $d=32$, and let $Y_n:\Omega_n\rightarrow[0,1]$ be the grayscale intensity. For direction $\delta\in\{\mathrm{h},\mathrm{v}\}$, let $\Omega_n^{\delta}$ contain pixels with a valid forward neighbor and let $\nabla_\delta$ be the corresponding finite difference. We reblur $Y_n$ along direction $\delta$ with a one-dimensional mean kernel of length $L_b=9$, producing $B_n^\delta$. The original variation, reblurred variation, and nonnegative attenuation are
\begin{align}
\Delta_{Y,n}^{\delta}(u)&=|\nabla_{\delta}Y_n(u)|,\\
\Delta_{B,n}^{\delta}(u)&=|\nabla_{\delta}B_n^{\delta}(u)|,\\
\Xi_n^{\delta}(u)&=\max\!\left(0,
\Delta_{Y,n}^{\delta}(u)-\Delta_{B,n}^{\delta}(u)\right).
\end{align}
These pixelwise blur quantities are distinct from the frame-level visual-change cue $\Delta_n$ in Equation~\eqref{eq:visual-change}. Following the no-reference perceptual blur metric of \citet{crete2007blur}, the directional blur estimate and frame sharpness are
\begin{align}
\beta_n^{\delta}
&=\frac{
\sum_{u\in\Omega_n^{\delta}}\Delta_{Y,n}^{\delta}(u)
-\sum_{u\in\Omega_n^{\delta}}\Xi_n^{\delta}(u)}
{\sum_{u\in\Omega_n^{\delta}}\Delta_{Y,n}^{\delta}(u)},\\
s_n^{\mathrm{sharp}}
&=1-\max_{\delta\in\{\mathrm{h},\mathrm{v}\}}\beta_n^{\delta}.
\end{align}
If a direction has no measurable original variation, its denominator is zero and we set $\beta_n^\delta=1$. Thus a flat image is not treated as sharp merely because reblurring leaves it unchanged.

For exposure, let $I_{n,c}(u)\in[0,1]$ denote color channel $c$ at pixel $u$ of the resized color thumbnail. Following the well-exposedness term of \citet{mertens2007exposure}, we compute
\begin{align}
W_n(u)&=\prod_{c\in\{R,G,B\}}
\exp\!\left(-\frac{(I_{n,c}(u)-\mu)^2}{2\sigma^2}\right),\\
\ell_n^{\mathrm{light}}
&=\frac{1}{|\Omega_n|}\sum_{u\in\Omega_n}W_n(u),
\end{align}
with $\mu=0.5$ and $\sigma=0.2$. The channel product downweights pixels with any channel far from mid-range. We average over pixels and combine the resulting exposure score with sharpness:
\begin{equation}
O_n=s_n^{\mathrm{sharp}}\ell_n^{\mathrm{light}}.
\end{equation}
Observability contributes only through $\lambda_oO_n$ in Equation~\eqref{eq:evidence}; it does not impose a hard filtering threshold. This term downweights severely blurred or poorly exposed frames but does not measure query relevance or semantic informativeness.

\subsection{Greedy Matryoshka Ranking Algorithm}
\label{sec:appendix-greedy-ranking}

Algorithm~\ref{alg:greedy-matryoshka} implements the position-dependent surrogate from Section~3.4. Starting from the budget-independent candidate pool, it appends one frame at a time without replacing earlier selections. Each intermediate sequence is therefore an exact prefix of the final ranking.

\begin{algorithm}[htbp]
\caption{Greedy construction of the Matryoshka ranking}
\label{alg:greedy-matryoshka}
\small
\textbf{Input}: $\mathcal{C}(q)$ with $E_n(q)$ and $\mathbf f_n$; video length $N$\\
\textbf{Parameters}: $M,w_e^{+},w_e^{-},w_d,\gamma$\\
\textbf{Output}: Shared ranking $\mathcal{R}$
\begin{algorithmic}[1]
\STATE $r_1\leftarrow\arg\max_{n\in\mathcal{C}(q)}E_n(q)$
\STATE $\mathcal{R}\leftarrow(r_1)$; $\mathcal{S}\leftarrow\{r_1\}$
\FOR{$k=2,\ldots,M$}
    \STATE Compute $w_e(k)$ using Equation~\eqref{eq:weights}
    \STATE $w_c(k)\leftarrow1-w_e(k)-w_d$
    \FOR{each $n\in\mathcal{C}(q)\setminus\mathcal{S}$}
        \STATE Compute $H_n(k)$ and $D_n(k)$ using Equations~\eqref{eq:coverage}--\eqref{eq:diversity}
        \STATE $u_k(n)\leftarrow w_e(k)E_n(q)+w_c(k)H_n(k)+w_dD_n(k)$
    \ENDFOR
    \STATE $r_k\leftarrow\arg\max_{n\in\mathcal{C}(q)\setminus\mathcal{S}}u_k(n)$
    \STATE Append $r_k$ to $\mathcal{R}$
    \STATE $\mathcal{S}\leftarrow\mathcal{S}\cup\{r_k\}$
\ENDFOR
\STATE \textbf{return} $\mathcal{R}$
\end{algorithmic}
\end{algorithm}

For a budget $K\leq M$, we take the prefix $\mathcal{R}_{:K}$ and sort its frame indices by timestamp before passing the frames to the LMM. No additional selection run is needed.

\section{Limitations}
\label{sec:appendix-limitations}

\paragraph{Sparse-discovery blind spots.}
\textsc{MEC} scores a duration-adaptive sparse grid, activates the highest-scoring temporal segments, and samples around one anchor per active segment. This limits computation but can miss brief events between probes. Segment averaging may further dilute weak evidence at neighboring probes and leave a relevant segment inactive. Because ranking operates only on the discovered candidates, subsequent coverage and diversity terms cannot recover omitted frames. This is particularly problematic for very long videos, single-frame state changes, or events with little visual continuity, as illustrated in Figure~\ref{fig:failure-case}. Denser probes, shorter segments, higher activation ratios, or multiple anchors could improve recall at the cost of more decoding and query scoring. Adaptive boundary proposals or refinement triggered by uncertainty are possible alternatives.

\paragraph{Approximate evidence cues.}
The evidence cues approximate downstream utility. BLIP-2 ITM scores frames individually and may miss evidence that depends on ordered observations or long temporal relationships. The descriptor with a 32-pixel short side discards color and fine detail; cuts, camera motion, and lighting changes can dominate its distances even when small semantic changes matter more. Sharpness and exposure can also downweight an indispensable frame that is blurred or poorly lit. These errors affect segment activation and early ranks, and the fixed evidence mixture cannot adapt its cue weights to each question. Temporal scorers, higher-resolution descriptors, or learned calibration could help, but may require additional training, memory, or per-query computation.

\paragraph{Greedy surrogate rather than global optimization.}
Equation~\eqref{eq:matryoshka-objective} defines an oracle objective over all prefixes. Prefix exposure motivates position-dependent weights, but evidence, coverage, and diversity remain proxies for unknown downstream marginal utility. Greedy construction guarantees nesting, not global optimality at any budget or for their weighted aggregate. An individually relevant frame can enter early and remain selected even if another combination would be more useful. The smooth exposure profile is a design prior, not a fit to deployment weights $\pi_K$; deployments concentrated on a few budgets may benefit from a different transition. The ablations provide empirical support for the chosen schedule. Beam search, local swaps, or learned marginal-utility estimates could improve selection at additional computational cost.

\section{Additional Experiments}
\label{sec:appendix-additional-experiments}

\subsection{Additional Ranking Ablations}

\begin{table}[htbp]
\caption{Additional ablations on Video-MME Overall with Qwen3.5-9B. Budget groups follow Table~\ref{tab:ablation}; Uniform and the default \textsc{MEC} configuration are repeated for reference. Arrows give the start and end evidence weights of linear schedules, with $w_d=.2$ and $w_c(k)=1-w_e(k)-w_d$. Best values are bold.}
\label{tab:appendix-ranking-ablation}
\centering
\small
\setlength{\tabcolsep}{5pt}
\renewcommand{\arraystretch}{1.06}
\begin{tabularx}{0.8\linewidth}{@{}>{\raggedright\arraybackslash}Xrrrr@{}}
\toprule
Variant & Low & Med. & High & Avg. \\
\midrule
Uniform & 60.80 & 67.81 & 73.09 & 67.23 \\
\oursrow \textbf{\textsc{MEC}} (default) & \textbf{65.19} & 70.46 & 74.19 & \textbf{69.94} \\
Evidence only & 61.37 & 67.54 & 73.19 & 67.36 \\
Linear $.2\!\to\!.6$ & 63.19 & 69.52 & 74.28 & 69.00 \\
Linear $.7\!\to\!.3$ & 63.80 & 70.17 & \textbf{74.48} & 69.48 \\
Linear $.5\!\to\!.1$ & 65.00 & \textbf{70.59} & 73.67 & 69.75 \\
\bottomrule
\end{tabularx}
\end{table}

Table~\ref{tab:appendix-ranking-ablation} extends the comparisons in Table~\ref{tab:ablation}. Combining evidence, coverage, and diversity improves average accuracy by $2.58$ pp over Evidence only. With candidate discovery fixed, reversing the default evidence-to-context transition reduces the average by $0.94$ pp. Changing the linear endpoints from $.6\!\to\!.2$ to $.7\!\to\!.3$ or $.5\!\to\!.1$ gives averages of $69.48$ and $69.75$. Together with the three decay rates in Table~\ref{tab:ablation-schedules}, these five evidence-decay schedules fall within $0.46$ pp of one another and outperform Uniform by at least $2.25$ pp. The gains hold across the tested decay rates and endpoints.

\subsection{Generalization Across Qwen3.5 Model Scales}

\begin{table}[htbp]
\caption{Video-MME Overall accuracy across Qwen3.5 model scales and frame budgets. All models use the same configuration.}
\label{tab:appendix-model-scales}
\centering
\small
\begingroup
\setlength{\tabcolsep}{3pt}
\renewcommand{\arraystretch}{1.04}
\begin{tabularx}{0.8\columnwidth}{@{}ll*{6}{>{\centering\arraybackslash}X}@{}}
\toprule
Size & Selector & 8 & 16 & 32 & 64 & 128 & 256 \\
\midrule
4B & Uniform & 56.19 & 60.04 & 62.93 & 65.30 & 67.15 & 68.70 \\
\oursrow 4B & \textbf{\textsc{MEC}} & \textbf{60.52} & \textbf{62.74} & \textbf{65.89} & \textbf{67.04} & \textbf{68.59} & \textbf{69.37} \\
\midrule
9B & Uniform & 58.96 & 62.63 & 66.15 & 69.48 & 72.48 & 73.70 \\
\oursrow 9B & \textbf{\textsc{MEC}} & \textbf{63.85} & \textbf{66.52} & \textbf{69.11} & \textbf{71.81} & \textbf{73.74} & \textbf{74.63} \\
\midrule
35B-A3B & Uniform & 61.85 & 65.44 & 69.33 & 72.67 & 74.93 & 76.59 \\
\oursrow 35B-A3B & \textbf{\textsc{MEC}} & \textbf{65.59} & \textbf{69.11} & \textbf{72.56} & \textbf{75.37} & \textbf{76.07} & \textbf{77.30} \\
\bottomrule
\end{tabularx}
\endgroup

\end{table}

Table~\ref{tab:appendix-model-scales} compares frame selection across downstream model scales. The six-budget mean under Uniform rises from $63.39$ for 4B to $67.23$ for 9B and $70.14$ for 35B-A3B. The corresponding \textsc{MEC} means are $65.69$, $69.94$, and $72.67$. Larger models achieve higher accuracy with either selector, while \textsc{MEC} retains an advantage at every scale.

\textsc{MEC} improves all 18 model--budget pairs. The mean gains are $2.31$, $2.71$, and $2.53$ pp for 4B, 9B, and 35B-A3B, respectively, rising to $4.33$, $4.89$, and $3.74$ pp at $K=8$. Thus, selecting relevant evidence remains useful even for the larger model under a small frame budget. The 35B-A3B model also gains $3.23$ and $2.70$ pp at $K=32$ and $64$, respectively.

At $K=256$, the gains narrow to $0.67$, $0.93$, and $0.71$ pp as both selectors cover more of the video. The positive margins are consistent with the intended ranking behavior: later frames expand context while shorter-prefix evidence remains selected.

\subsection{Thinking-Mode Results}
\label{sec:appendix-thinking}

Our standard evaluations disable Qwen3.5 thinking through the official interface \citep{qwen2026qwen35}, as described in Section~\ref{sec:appendix-downstream-models}. Figure~\ref{fig:fine-grained-budget} (right) enables thinking for both Uniform and \textsc{MEC}. The gains over Uniform are $5.56$, $4.11$, $2.78$, $2.29$, and $2.04$ pp at $K=8$, $16$, $32$, $64$, and $128$, respectively. They decrease as the budget grows but remain positive at every tested budget.

\subsection{Additional Qualitative Example}

\begin{figure}[htbp]
\centering
\includegraphics[width=0.65\columnwidth]{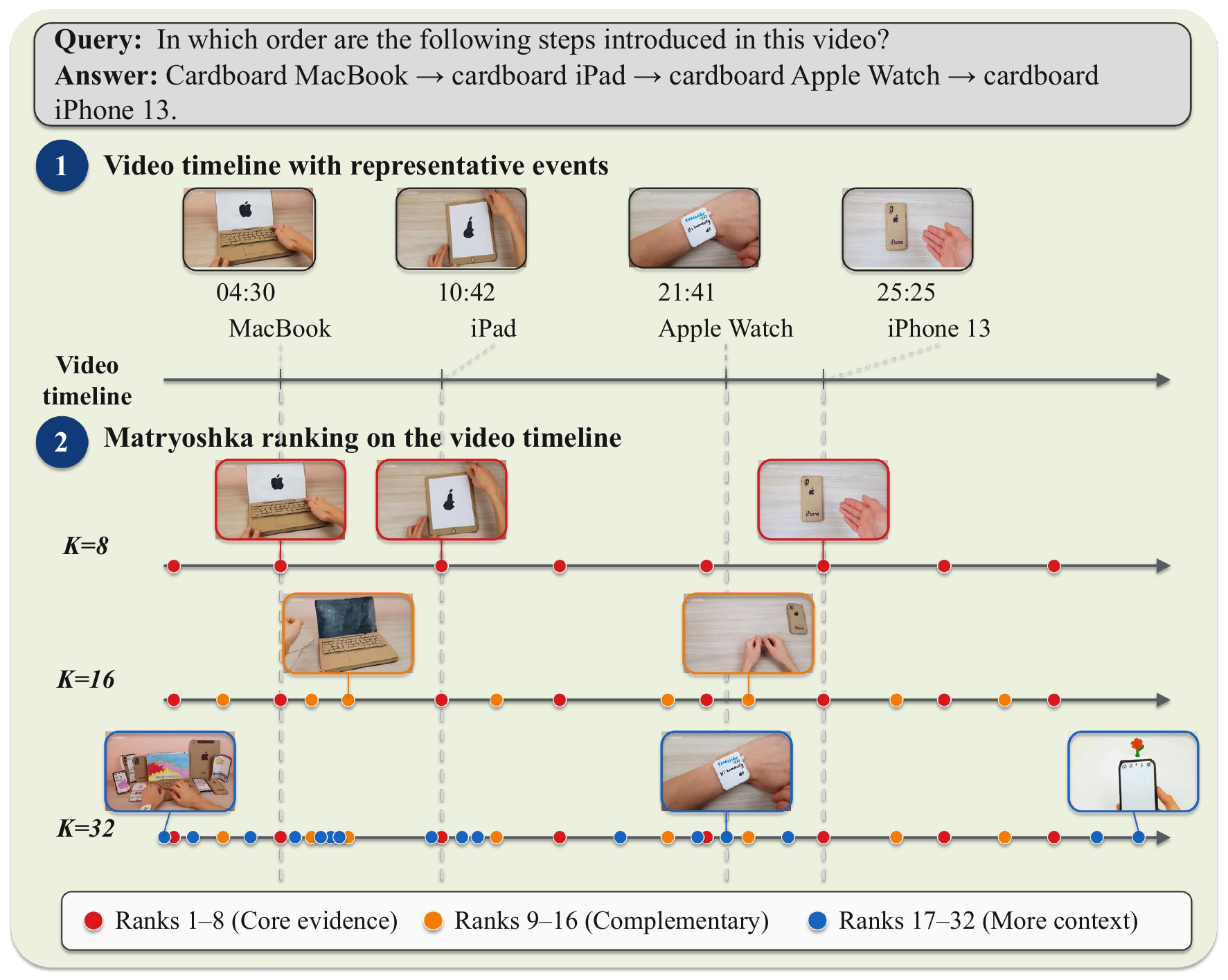}
\caption{Ordering four cardboard-device introductions: MacBook (04:30), iPad (10:42), Apple Watch (21:41), and iPhone~13 (25:25). Red, orange, and blue mark frames entering at ranks 1--8, 9--16, and 17--32. Each prefix is shown in the chronological order used for LMM inference.}
\label{fig:appendix-success-case}
\end{figure}

Figure~\ref{fig:appendix-success-case} shows a query about four introductions spread across a video. The eight-frame prefix covers all four rather than concentrating on one construction sequence. In chronological order, these frames show the MacBook, iPad, Apple Watch, and iPhone introductions, and the LMM answers correctly. The $K=16$ prefix retains these frames and adds MacBook and iPhone construction views. At $K=32$, additions include an opening overview, another Apple Watch view, and a later iPhone demonstration. The nested prefixes extend coverage while preserving earlier selections, although this example does not establish that every added frame is needed for the answer.

\clearpage
\end{document}